\documentclass[11pt]{article}

\usepackage[preprint]{acl}

\usepackage{times}
\usepackage{latexsym}

\usepackage{graphicx}
\usepackage{wrapfig}
\usepackage{amssymb}
\usepackage{amsmath}
\usepackage{bm}
\usepackage{caption}
\usepackage{multirow}
\usepackage{mathrsfs}
\usepackage{mathtools}
\usepackage{algorithm}
\usepackage{algpseudocode}
\usepackage{soul}
\usepackage{booktabs}

\usepackage[T1]{fontenc}

\usepackage[utf8]{inputenc}

\usepackage{microtype}

\usepackage{inconsolata}

\usepackage{graphicx}
\usepackage{hyperref}

\title{CORRECT: Context- and Reference-Augmented Reasoning and Prompting for Fact-Checking}

\author{Delvin Ce Zhang \\
  The Pennsylvania State University \\
  \texttt{delvincezhang@gmail.com} \\\And
  Dongwon Lee \\
  The Pennsylvania State University \\
  \texttt{dongwon@psu.edu} \\}

\begin{document}
\maketitle
\begin{abstract}
Fact-checking the truthfulness of claims usually requires reasoning over multiple evidence sentences. Oftentimes, evidence sentences may not be always self-contained, and may require additional contexts and references from elsewhere to understand coreferential expressions, acronyms, and the scope of a reported finding. For example, evidence sentences from an academic paper may need contextual sentences in the paper and descriptions in its cited papers to determine the scope of a research discovery. However, most fact-checking models mainly focus on the reasoning within evidence sentences, and ignore the auxiliary contexts and references. To address this problem, we propose a novel method, Context- and Reference-augmented Reasoning and Prompting. For evidence reasoning, we construct a three-layer evidence graph with evidence, context, and reference layers. We design intra- and cross-layer reasoning to integrate three graph layers into a unified evidence embedding. For verdict prediction, we design evidence-conditioned prompt encoder, which produces unique prompt embeddings for each claim. These evidence-conditioned prompt embeddings and claims are unified for fact-checking. Experiments verify the strength of our model. Code and datasets are available at \href{https://github.com/cezhang01/correct}{https://github.com/cezhang01/correct}.
\end{abstract}

\section{Introduction}

\begin{figure}[t]
	\centering
	\includegraphics[width=1\linewidth]{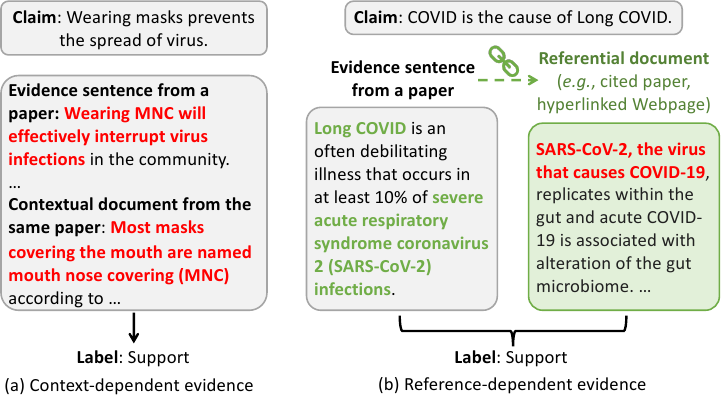}
	\caption{Illustration of (a) context-dependent and (b) reference-dependent evidence from BearFact dataset.}
	\label{fig:illustration}
\end{figure}

The proliferation of misinformation has posed growing challenge in the realm of information reliability. There is a need to develop automated fact-checking methods \cite{survey} to verify the truthfulness of real-world claims using evidence.

Existing fact-checking models \cite{gear,kgat} have shown promise in aggregating and reasoning over multiple evidence sentences to verify a claim. However, the evidence sentences retrieved from a large corpus may contain incomplete information when they are taken out-of-corpus. We need to refer to additional contexts and references from elsewhere to understand coreferential expressions, acronyms, and the scope of a reported finding. For example, Fig. \ref{fig:illustration}(a) illustrates context-dependent evidence, where undefined acronym ``MNC'' in evidence sentence from a paper abstract requires additional context from the abstract to jointly interpret the meaning of acronym ``MNC''. Fig. \ref{fig:illustration}(b) presents reference-dependent evidence, where we need to check the cited paper to understand that ``SARS-CoV-2 infection'' and ``COVID-19 infection'' are coreferential expressions, so that we could accurately fact-check the claim. Such scenario also exists in general domain where evidence sentences from a Wikipedia page may need contextual sentences in the same page and text in the hyperlinked pages to complement the insufficient information in the evidence.

\textbf{Challenges and Approach.} To overcome the limitations of existing methods, we propose \ul{\textbf{Co}}ntext- and \ul{\textbf{R}}eference-augmented \ul{\textbf{Re}}asoning and prompting for fa\ul{\textbf{ct}}-checking (CORRECT), to address two open questions.

First, \emph{how to aggregate both contextual and referential documents into evidence reasoning?} Some models are proposed to capture contextual documents, e.g., MultiVerS \cite{multivers}. Some others are designed for referential documents, e.g., Transformer-XH \cite{transformer_xh} and HESM \cite{hesm}. However, they incorporate either contextual or referential documents, failing to aggregate both of them into unified evidence embedding. Moreover, most of them simply concatenate evidence with contextual or referential documents, and inefficiently input the long text to language models for evidence encoding. Though they have shown that modeling either contexts or references helps fact-checking, integrating both of them for evidence reasoning is still unexplored. In our model, we construct a three-layer graph with evidence, context, and reference layers. We design intra- and cross-layer reasoning to aggregate three graph layers into unified evidence embedding.

Second, \emph{how to integrate evidence reasoning and claim for accurate verdict prediction?} Previous fact-checking methods, e.g., ProToCo \cite{protoco}, rely on natural language as input prompt to language model for claim verification. However, discrete natural language prompts are difficult to design and may result in suboptimal results \cite{coop}. Recently, prompt tuning \cite{soft_prompting} uses continuous and learnable prompt embeddings to replace discrete prompt and has achieved decent result, but no one has explored its design for claim verification. We propose evidence-conditioned prompt encoder, which takes evidence embedding as input, and produces unique prompt embeddings for each claim. We combine prompt embeddings with claim token embeddings to unify evidence and claim for verdict prediction.

\textbf{Contributions.} First, we propose a novel model, Context- and Reference-augmented Reasoning and Prompting (CORRECT), to integrate both contextual and referential documents into evidence reasoning. Second, we design a three-layer evidence graph, and propose intra- and cross-layer reasoning to learn unified evidence embedding. Third, we propose evidence-conditioned prompt embeddings, which are combined with claims to integrate evidence reasoning with claim for fact-checking.
\section{Related Work}

\textbf{Multi-hop fact-checking.} Complex claims usually require reasoning over multiple evidence sentences. Many methods are based on Language Models \cite{transformer,bert} and Graph Neural Networks \cite{graphsage}, such as GEAR \cite{gear}, KGAT \cite{kgat}, DREAM \cite{dream}, SaGP \cite{sagp}, DECKER \cite{decker}, CausalWalk \cite{causalwalk}, etc. However, they mainly focus on the reasoning within evidence sentences. They ignore the auxiliary contextual and referential documents. Methods incorporating contextual documents are proposed, e.g., ParagraphJoint \cite{paragraphjoint}, ARSJoint \cite{arsjoint}, MultiVerS \cite{multivers}, etc. Some others integrating referential documents include Transformer-XH \cite{transformer_xh} and HESM \cite{hesm}. However, they incorporate either contextual or referential documents, but not both. In contrast, we construct a three-layer evidence graph to model evidence sentences, contexts, and references. There are fake news detection models where auxiliary graph with Wikidata is used \cite{fake_news_detection,fake_news_detection2}. Fake news detection aims to detect the whole article with meta-data, while fact-checking focuses on claim sentences with retrieved evidence.

Some fact-checking works are based on retrieval-augmented generation \cite{justilm}. They unify evidence retrieval and claim verification as a joint approach, while our model mainly focuses on verification, and relies on external tool for evidence retrieval. Our setting is consistent with existing works \cite{multivers,causalwalk}.

\textbf{Prompt-based fact-checking.} Some models verify claims by prompting LLMs \cite{gpt4}. ProToCo \cite{protoco} inputs both evidence sentences and claim to T5 \cite{t5}. ProgramFC \cite{programfc} decomposes complex claims into simpler sub-tasks and uses natural language to prompt LLMs. Varifocal \cite{varifocal} formulates fact-checking as question generation and answering. They rely on handcrafted natural language as prompt. The performance heavily relies on the choice of prompt, and it is difficult to design a prompt that produces a decent result, as shown in \cite{coop}. Our model is designed with learnable prompt embeddings where the prompting instruction is naturally learned by embeddings through optimization.

\textbf{Prompt learning.} Prompting \cite{prompting} uses natural language as the input to language models to fulfill certain tasks. Many prompting models have been proposed, including natural language prompt \cite{discrete_prompting,discrete_prompting2} and prompt embeddings \cite{soft_prompting,p_tuning,p_tuning_v2,prefix_tuning}. Prompting also benefits many tasks \cite{cocoop,machine_translation}. However, no one has explored prompt embeddings for fact-checking.

\textbf{Text-attributed graph.} Texts are usually connected in a graph structure, termed text-attributed graph \cite{tag_survey}. Various methods have been developed to learn text embeddings in an unsupervised manner \cite{adjacent_encoder,dbn,semivn,hgtm,graphformers,patton,hypformer}. Though both our model and these works construct a text-attributed graph, our work is different from them, since our model is a supervised model for fact-checking.
\section{Model Architecture}

\begin{table}[t]
	\centering
	\caption{Summary of mathematical notations.}
	\resizebox{\columnwidth}{!}{
		\begin{tabular}{c|l}
			\toprule
			Notation  & Description  \\
			\hline
			$ \mathcal{D} $ & a fact-checking dataset \\
			$ \mathcal{X} $ & a set of $ N=|\mathcal{X}| $ claims \\
			$ \mathcal{E} $ & a corpus of evidence sentences \\
			$ \mathcal{C} $ & a set of contextual documents \\
                $ \mathcal{R} $ & a set of referential documents \\
			$ \mathcal{N}_{\text{ref}}(e) $ & evidence sentence $ e $'s referential documents \\
                $ \mathcal{N}_{\text{evid}}(x) $ & claim $ x $'s retrieved evidence sentences \\
                $ \mathcal{Y} $ & a set of labels \\
                $ \textbf{h}_{e,\text{CLS}}^{(l)} $ & evidence sentence $ e $'s [CLS] token embedding \\
                $ \hat{\textbf{h}}_{c}^{(l)} $ & aggregated contextual document embedding \\
                $ \hat{\textbf{h}}_{r}^{(l)} $ & aggregated referential document embedding \\
                $ \widehat{\textbf{H}}_{e}^{(l)} $ & evidence sentence $ e $'s augmented embedding matrix \\
                $ \bm{\pi}_{m,y} $ & the $ m $-th prompt embedding for class $ y $ \\
                $ \textbf{h}_{m,y} $ & the $ m $-th base prompt embedding for class $ y $ \\
			\bottomrule
		\end{tabular}
	}
	\label{table:notation_table}
\end{table}

\begin{figure*}[t]
	\centering
	\includegraphics[width=1\linewidth]{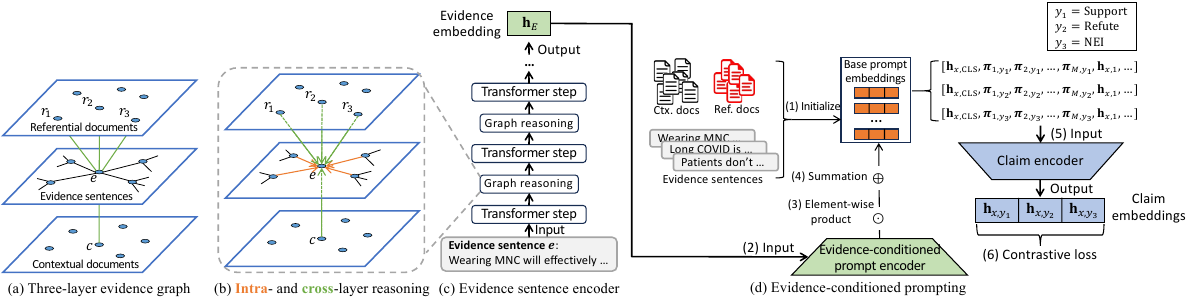}
	\caption{Model architecture. (a) A three-layer graph for a claim. (b) Intra- and cross-layer reasoning. (c) A nested architecture with language model and graph reasoning for evidence encoding. (d) Evidence-conditioned prompting.}
	\label{fig:model}
\end{figure*}

We introduce \ul{\textbf{Co}}ntext- and \ul{\textbf{R}}eference-augmented \ul{\textbf{Re}}asoning and prompting for fa\ul{\textbf{ct}}-checking (CORRECT). Table \ref{table:notation_table} summarizes math notations.

\subsection{Problem Formulation}

We are given a fact-checking dataset $ \mathcal{D}=\{\mathcal{X},\mathcal{E},\mathcal{C},\mathcal{R}\} $. Claim set $ \mathcal{X}=\{x_i\}_{i=1}^N $ contains a set of $ N $ claims. Evidence set $ \mathcal{E}=\{e_j\}_{j=1}^E $ is a corpus of $ E $ evidence sentences. For each evidence sentence $ e\in\mathcal{E} $, we have its contextual document $ c\in\mathcal{C} $. Usually, an evidence sentence has only one contextual document, from which this sentence is retrieved. We also have $ e $'s referential documents $ \mathcal{N}_{\text{ref}}(e)=\{r_{e,n}\}_{n=1}^{R_e}\subset\mathcal{R} $. Here $ R_e $ is the number of $ e $'s referential documents. Evidence sentence $ e $ may have multiple referential documents, such as papers cited by $ e $'s paper or Webpages hyperlinked by $ e $. We use $ \mathcal{N}_{\text{ref}}(e) $ to represent the set of $ e $'s referential documents. We use $ \mathcal{N}_{\text{evid}}(x)\subset\mathcal{E} $ to denote the set of evidence sentences for a claim $ x $.

Given $ \mathcal{D} $ as input, we design a model that uses evidence sentences from $ \mathcal{E} $ together with their contextual documents in $ \mathcal{C} $ and referential documents in $ \mathcal{R} $ to verify claims. Eventually, for each claim $ x\in\mathcal{X} $, we output its predicted label $ \hat{y}\in\mathcal{Y}=\{\text{SUPPORT, REFUTE, NEI}\} $, indicating whether the evidence supports, refutes, or does not have enough information to verify the claim.

As shown in Fig. \ref{fig:model}, CORRECT has two modules: (a-c) context- and reference-augmented evidence reasoning on three-layer graph, (d) evidence-conditioned prompting for claim verification.

\subsection{Three-layer Evidence Graph Reasoning}

\textbf{Graph construction.} For each claim $ x\in\mathcal{X}$ and its evidence sentences $ \mathcal{N}_{\text{evid}}(x)\subset\mathcal{E} $, we construct a three-layer graph with evidence, context, and reference layers in Fig. \ref{fig:model}(a). We consider evidence sentences, contextual documents, and referential documents as three types of vertices. Each type of vertices reside on their own layer. Cross-layer links between evidence layer and context layer connect each evidence sentence with its contextual document. Each evidence sentence and its referential documents are connected by cross-layer referential links. Green links in Fig. \ref{fig:model}(a) are cross-layer links. For multi-evidence reasoning, we add intra-layer links on evidence layer where evidence sentences of a claim are fully connected, shown by black links in Fig. \ref{fig:model}(a). The purpose of constructing three layers instead of mixing all vertices into one layer is to better differentiate three types of vertices.

\textbf{Intra-layer reasoning.} Evidence reasoning includes intra- and cross-layer reasoning. We first show intra reasoning (orange arrows in Fig. \ref{fig:model}(b)).

For each evidence sentence $ e\in\mathcal{N}_{\text{evid}}(x) $, we let $ \textbf{H}_e^{(l)}=[\textbf{h}_{e,\text{CLS}}^{(l)},\textbf{h}_{e,1}^{(l)},\textbf{h}_{e,2}^{(l)},...] $ denote the output from the $ l $-th Transformer step. Note that previous works call it the $ l $-th layer, but to distinguish it from our three-layer graph, we instead call it the $ l $-th step. $ \textbf{h}_{e,i}^{(l)}\in\Bbb R^d $ is $ d $-dimensional token embedding. We use graph neural network to aggregate different evidence sentences of a claim. For each evidence sentence $ e $, we first project it by
\begin{equation}
\label{eq:linear_projection}
    \tilde{\textbf{h}}_{e,\text{CLS}}^{(l)}=\textbf{W}_1\textbf{h}_{e,\text{CLS}}^{(l)}.
\end{equation}
The [CLS] token is taken as the evidence sentence embedding, and $ \textbf{W}_1\in\Bbb R^{d\times d} $ is type-specific parameter. We design type-specific attention.
\begin{equation}
\label{eq:neighbor_attention}
\resizebox{\columnwidth}{!}{
	$ a_{e,e^\prime}=\text{softmax}\Bigl(\text{LeakyReLU}(\textbf{b}_1^{\top}[\tilde{\textbf{h}}_{e,\text{CLS}}^{(l)}||\tilde{\textbf{h}}_{e^\prime,\text{CLS}}^{(l)}])\Bigl). $
 }
\end{equation}
$ e^\prime\in\mathcal{N}_{\text{evid}}(x)\textbackslash e $ is another evidence sentence for the same claim $ x $, $ [\cdot||\cdot] $ is concatenation, and $ \textbf{b}_1\in\Bbb R^{2d} $ is learnable parameter. Finally, we aggregate evidence sentences to $ e $ by mean pooling.
\begin{equation}
\label{eq:neighbor_aggregation}
    \hat{\textbf{h}}_e^{(l)}=\text{mean}\Bigl(\tilde{\textbf{h}}_{e,\text{CLS}}^{(l)},\sum_{e^\prime \in \mathcal{N}_{\text{evid}}(x)\textbackslash e}a_{e,e^\prime}\tilde{\textbf{h}}_{e^\prime,\text{CLS}}^{(l)}\Bigl).
\end{equation}
The aggregated sentence embedding $ \hat{\textbf{h}}_e^{(l)} $ captures information of both itself and other evidence sentences. To summarize Eqs. \ref{eq:linear_projection}--\ref{eq:neighbor_aggregation}, we have
\begin{equation}
\label{eq:graph_conv_layer}
\resizebox{\columnwidth}{!}{
    $ \hat{\textbf{h}}_e^{(l)}=f_{\text{GNN}}\Big(\textbf{h}_{e,\text{CLS}}^{(l)},\{\textbf{h}_{e^\prime,\text{CLS}}^{(l)}|e^\prime\in\mathcal{N}_{\text{evid}}(x)\textbackslash e\};\textbf{W}_1,\textbf{b}_1\Big). $
}
\end{equation}

To integrate intra-layer aggregation into the encoding of each evidence sentence, we introduce a \emph{virtual token} to represent the aggregated sentence embedding $ \hat{\textbf{h}}_e^{(l)} $. 
For evidence sentence $ e $, we concatenate $ \hat{\textbf{h}}_e^{(l)} $ with $ e $'s text token embeddings by $ \widehat{\textbf{H}}_e^{(l)}=\hat{\textbf{h}}_e^{(l)}||\textbf{H}_e^{(l)} $.
After concatenation, $ \widehat{\textbf{H}}_e^{(l)} $ contains information of both evidence sentence $ e $'s text and the aggregated embedding from $ e $'s intra-layer neighbors. We aim to propagate the aggregated sentence embedding to other text tokens of sentence $ e $, so that the text tokens can fully unify other sentences for multi-evidence reasoning. We will introduce \emph{asymmetric} multi-head self-attention to achieve this goal. But before that, we first discuss cross-layer reasoning.

\textbf{Cross-layer reasoning.} We present cross-layer reasoning, which aggregates contextual and referential documents into evidence sentences (green arrows in Fig. \ref{fig:model}(b)). The aggregation from referential documents to evidence sentence is similarly defined by Eq. \ref{eq:graph_conv_layer_ref}. We use reference-specific parameters, $ \textbf{W}_2 $ and $ \textbf{b}_2 $, to preserve graph heterogeneity.
\begin{equation}
\label{eq:graph_conv_layer_ref}
\resizebox{\columnwidth}{!}{
    $ \hat{\textbf{h}}_r^{(l)}=f_{\text{GNN}}\Big(\textbf{h}_{e,\text{CLS}}^{(l)},\{\textbf{h}_{r,\text{CLS}}^{(l)}|r\in\mathcal{N}_{\text{ref}}(e)\};\textbf{W}_2,\textbf{b}_2\Big). $
}
\end{equation}
Each referential document $ r\in\mathcal{N}_{\text{ref}}(e) $ is also encoded, and its [CLS] token is passed to Eq. \ref{eq:graph_conv_layer_ref} for aggregation. Similarly, we have $ \hat{\textbf{h}}_c^{(l)} $ as contextual document embedding. To integrate both embeddings into evidence sentence for cross-layer reasoning, we introduce two more \emph{virtual tokens}.
\begin{equation}
\label{eq:virtual_token}
    \widehat{\textbf{H}}_e^{(l)}=\hat{\textbf{h}}_c^{(l)}||\hat{\textbf{h}}_r^{(l)}||\hat{\textbf{h}}_e^{(l)}||\textbf{H}_e^{(l)}.
\end{equation}
The augmented embedding matrix, i.e., $ \widehat{\textbf{H}}_e^{(l)} $, contains both intra-evidence reasoning as well as cross-layer context and reference augmentation.

To fully unify all three graph layers into evidence sentence $ e $, we input $ \widehat{\textbf{H}}_e^{(l)} $ at Eq. \ref{eq:virtual_token} to the $ (l+1) $-th Transformer step with our proposed \emph{asymmetric} multi-head self-attention ($ \text{MSA}^{\text{asy}} $).
\begin{equation}
\label{eq:asymmetric_attention}
\resizebox{\columnwidth}{!}{
$ \begin{split}
    \text{MSA}^{\text{asy}}(\textbf{H}_e^{(l)},\widehat{\textbf{H}}_e^{(l)},&\widehat{\textbf{H}}_e^{(l)})=\text{softmax}\Big(\dfrac{\textbf{Q}\textbf{K}^\top}{\sqrt{d}}\Big)\textbf{V},\\
    \textbf{Q}=\textbf{H}_e^{(l)}\textbf{W}_{Q}^{(l)},\;\  \textbf{K}&=\widehat{\textbf{H}}_e^{(l)}\textbf{W}_{K}^{(l)},\;\  \textbf{V}=\widehat{\textbf{H}}_e^{(l)}\textbf{W}_{V}^{(l)}.
\end{split} $
}
\end{equation}
Keys $ \textbf{K} $ and values $ \textbf{V} $ are augmented with virtual tokens, but queries $ \textbf{Q} $ are not, to avoid context and reference embeddings being overwritten by evidence sentence embedding. The result of asymmetric MSA is passed to a multi-layer perceptron and layer normalization \cite{transformer}. Finally, we obtain the output from the $ (l+1) $-th step, $ \textbf{H}_e^{(l+1)} $, integrating evidence sentence $ e $, other evidence sentences of the same claim, $ e $'s contextual and referential documents, see Fig. \ref{fig:model}(c).

We conduct such intra-layer and cross-layer reasoning inside each Transformer step to allow different graph layers to fully communicate with each other. We repeat such nested and graph-augmented encoding for $ L $ times, and obtain $ \textbf{h}_e=\textbf{h}_{e,\text{CLS}}^{(L)} $ as the graph-augmented embedding for evidence sentence $ e $. This nested architecture is shown by Fig. \ref{fig:model}(c). For claim $ x $, we have $ \{\textbf{h}_e\}_{e\in\mathcal{N}_{\text{evid}}(x)} $, a set of graph-augmented embeddings for its evidence sentences. Finally, we aggregate them by mean pooling and obtain a single evidence embedding.
\begin{equation}
\label{eq:evidence_embedding}
    \textbf{h}_E=\text{mean}(\textbf{h}_e|e\in\mathcal{N}_{\text{evid}}(x)).
\end{equation}

\subsection{Evidence-conditioned Prompting}

Now we integrate evidence reasoning into claim embedding to fully integrate their information for fact-checking. Prompting \cite{p_tuning} is a powerful method in fact-checking \cite{protoco}. However, existing models are mainly based on natural language as input prompt to language models for verdict prediction. Handcrafted discrete prompt has two disadvantages: First, it is difficult to manually design a prompt that provides a decent performance. Previous works \cite{coop} have shown that the change of a single word in the prompt may lead to significant deterioration of the results, and it is time-consuming to enumerate every prompt. Second, discrete natural language prompt is difficult to optimize, since language models are intrinsically continuous.

To mitigate these problems, we explore learnable and continuous prompt embedding. Below we design a prompt encoder, which takes evidence embedding $ \textbf{h}_E $ as input, and produces evidence-conditioned prompt embeddings. See Fig. \ref{fig:model}(d).

\textbf{Evidence-conditioned prompt encoder.} We consider below continuous embeddings as prompt.
\begin{equation}
\label{eq:input_embeddings_with_prompt}
    \textbf{P}_x=[\textbf{h}_{x,\text{CLS}},\bm{\pi}_1,\bm{\pi}_2,...,\bm{\pi}_M,\textbf{h}_{x,1},\textbf{h}_{x,2},...].
\end{equation}
Here $ \{\bm{\pi}_m\}_{m=1}^M $ where $ \bm{\pi}_m\in\Bbb R^d $ is a set of $ M $ learnable evidence-conditioned prompt embeddings to be explained shortly, and $ M $ is a hyperparameter, indicating the number of prompt embeddings. Each $ \textbf{h}_{x,i}\in\Bbb R^d $ is a $ d $-dimensional embedding of the $ i $-th text token in claim $ x $. In language models, there is an embedding look-up table before language model encoder. In this look-up table, input text tokens are first mapped to the vocabulary to obtain their token embeddings, which are then summed up with positional encodings. $ \textbf{h}_{x,i} $ in Eq. \ref{eq:input_embeddings_with_prompt} is obtained by this look-up table.

Now we explain prompt embeddings $ \{\bm{\pi}_m\}_{m=1}^M $, output from an evidence-conditioned prompt encoder. We first initialize $ M $ \emph{base} prompt embeddings, $ \{\textbf{h}_m\}_{m=1}^M $. We then project evidence embedding $ \textbf{h}_E $ in Eq. \ref{eq:evidence_embedding} to the prompt embedding space, followed by element-wise product and summation.
\begin{equation}
\label{eq:scaling_and_shifting}
\resizebox{\columnwidth}{!}{
    $ \bm{\alpha}_x=\text{tanh}\Big(\dfrac{\textbf{W}_\alpha\textbf{h}_E+\textbf{b}_\alpha}{\tau}\Big),\quad\bm{\beta}_x=\text{tanh}\Big(\dfrac{\textbf{W}_\beta\textbf{h}_E+\textbf{b}_\beta}{\tau}\Big), $
}
\end{equation}
\begin{equation}
\label{eq:scaling_and_shifting2}
    \bm{\pi}_m=\textbf{h}_m\odot(\bm{\alpha}_x+\textbf{1})+\bm{\beta}_x.
\end{equation}
$ \odot $ is element-wise product, and $ \textbf{1}\in\Bbb R^d $ is a vector of ones to ensure that the scaling of $ \textbf{h}_m $ is centered around one. $ \tau $ is a temperature to scale the shape of tanh function. $ \bm{\pi}_m $ is thus conditioned on evidence embedding, and different claims with their own evidence sentences should have their unique claim-specific prompt embeddings, shown by Fig. \ref{fig:model}(d).

Given the label set  $ \mathcal{Y}= $\{SUPPORT, REFUTE, NEI\}, which usually has three types of labels, we apply above evidence-conditioned prompt encoder and correspondingly obtain three sets of prompt embeddings, $ \{\bm{\pi}_{m,y}\}_{m=1}^M $ where $ y\in\mathcal{Y} $. As in Eq. \ref{eq:input_embeddings_with_prompt}, we concatenate each set of prompt embeddings with token embeddings of claim $ x $, and obtain three sets of inputs $ \{\textbf{P}_{x,y}\}_{y\in\mathcal{Y}} $ to claim encoder.
\begin{equation}
\resizebox{\columnwidth}{!}{
$ \begin{split}
    \textbf{H}_{x,y}^{(L)}&=f(\textbf{P}_{x,y})\\
    &=f([\textbf{h}_{x,\text{CLS}},\bm{\pi}_{1,y},\bm{\pi}_{2,y},...,\bm{\pi}_{M,y},\textbf{h}_{x,1},\textbf{h}_{x,2},...])
\end{split} $
}
\end{equation}
$ \textbf{H}_{x,y}^{(L)} $ is the output from the claim encoder, and its [CLS] token is taken as claim embedding $ \textbf{h}_{x,y}=\textbf{h}_{x,y,\text{CLS}}^{(L)} $. Claim encoder shares parameters with evidence encoder. Due to contextualized modeling, claim token embeddings and evidence-conditioned prompt embeddings fully exchange information, and the output claim embedding captures both claim $ x $ and evidence reasoning for fact-checking.

Finally, we use contrastive loss function to predict the veracity of claim $ x $ by
\begin{equation}
\label{eq:objective_function}
\resizebox{\columnwidth}{!}{
    $ \mathcal{L}=-\sum_{x\in\mathcal{X}_{\text{train}}}\log\dfrac{\exp(\textbf{h}_{x,y}^\top\textbf{h}_E)}{\exp(\textbf{h}_{x,y}^\top\textbf{h}_E)+\sum_{y^\prime\in\mathcal{Y}\textbackslash y}\exp(\textbf{h}_{x,y^\prime}^\top\textbf{h}_E)}. $
}
\end{equation}
$ \textbf{h}_E $ is evidence embedding of claim $ x $ obtained by Eq. \ref{eq:evidence_embedding}. $ \mathcal{X}_{\text{train}} $ is a set of training claims. Though we use three types of labels in $ \mathcal{Y} $, more types of labels in $ \mathcal{Y} $ can also be modeled. Algorithm \ref{algo:training_algorithm} summarizes the learning process.

\textbf{Initialization of base prompt embeddings.} Previous works \cite{coop} have shown the importance of the initialization of \emph{base} prompt embeddings $ \{\textbf{h}_{m,y}\}_{m=1}^M $ where $ y\in\mathcal{Y} $. Some of them randomly initialize the embeddings, while others use word embeddings of discrete prompts. Random initialization presents unstable optimization \cite{g2p2}, while it is difficult to choose the right discrete prompts for initialization. We solve these problems by using the three-layer graph.

For a claim $ x $, the vertices on its three-layer graph consistently carry the signal of claim $ x $'s veracity due to semantic relatedness. Thus, for each label in the label set $ y\in\mathcal{Y} $, we have training claims belonging to this label $ \mathcal{X}_{\text{train},y}=\{x\in\mathcal{X}_{\text{train}}|y_x=y\} $. For each of these claims, we truncate its evidence sentences, contextual and referential documents to $ M $ words, and obtain their $ M $ word embeddings in the look-up table of language model. We then take mean pooling for evidence sentences, contextual and referential documents, and obtain $ M $ pooled word embeddings for each claim. Finally, we average all training claims belonging to the same label $ \mathcal{X}_{\text{train},y} $, and obtain $ M $ word embeddings, which are used to initialize $ M $ base prompt embeddings $ \{\textbf{h}_{m,y}\}_{m=1}^M $. They are derived from training claims of the same label, thus provide a more informative starting point than random initialization for verdict prediction. We repeat this process for every label $ y\in\mathcal{Y} $, and obtain initialization for each set of base prompt embeddings.
\section{Experiments}

We conduct extensive experiments and ablation analysis to evaluate the effectiveness of the proposed model CORRECT.

\textbf{Datasets.} We use 4 datasets in Table \ref{table:dataset_statistics}. FEVEROUS \cite{feverous} is a general-domain dataset. Each claim is annotated in the form of sentences and/or cells from tables in Wikipedia pages. Since we focus on textual fact-checking, we follow \cite{programfc} and select claims that only require sentences as evidence. We call this subset \textbf{FEVEROUS-S}. \textbf{BearFact} \cite{bear_fact} is a biomedical dataset with sentences from papers as evidence. Its original dataset does not have evidence for claims in NEI class. We follow \cite{protoco} and select sentences that have the highest \textit{tf-idf} similarity with those claims as evidence. \textbf{Check-COVID} \cite{check_covid} contains claims about COVID-19. \textbf{SciFact} \cite{scifact} is a dataset with sentences in papers as evidence. As in its original paper, for claims in NEI class, we choose sentences from the cited abstract with top-3 highest \textit{tf-idf} similarity with the claim as evidence. Appendix \ref{sec:dataset_preprocessing} contains data preprocessing details.

\begin{table}
	\centering
	\caption{Dataset statistics.}
	\resizebox{\columnwidth}{!}{
		\begin{tabular}{c|cc|c|c}
			\toprule
			\multirow{2}{*}{Name} & \multicolumn{2}{c|}{\#Claims} & \#Contextual & \#Referential \\
			\cline{2-3}
			{} & Train & Test & Documents & Documents \\
			\hline
                FEVEROUS-S & 23,912 & 5,978 & 19,546 & 21,579 \\
			BearFact & 1,158 & 290 & 1,166 & 12,938 \\ 
			Check-COVID & 1,275 & 229 & 347 & 3,132 \\
			SciFact & 809 & 300 & 1,189 & 9,617 \\
			\bottomrule
		\end{tabular}
	}
	\label{table:dataset_statistics}
\end{table}

\begin{table*}[t]
	\centering
	\caption{Verdict prediction results on \emph{fully supervised} setting with \emph{Macro F1} score. Results are in percentage.}
	\vspace{-0.2cm}
	\resizebox{\textwidth}{!}{
		\begin{tabular}{c|cc|cc|cc|cc}
			\toprule
                \multirow{2}{*}{Model} & \multicolumn{2}{c|}{BearFact} & \multicolumn{2}{c|}{Check-COVID} & \multicolumn{2}{c|}{SciFact} & \multicolumn{2}{c}{FEVEROUS-S} \\
			\cline{2-9}
			{} & Gold & Retrieved & Gold & Retrieved & Gold & Retrieved & Gold & Retrieved \\
			\hline
                KGAT & 53.11$ \pm $2.25 & 36.55$ \pm $1.95 & 71.97$ \pm $1.31 & 75.83$ \pm $0.74 & 70.23$ \pm $1.08 & 59.83$ \pm $0.68 & 86.10$ \pm $0.32 & 67.76$ \pm $0.93 \\
                HESM & 44.90$ \pm $2.20 & 42.93$ \pm $0.27 & 62.85$ \pm $0.59 & 71.58$ \pm $1.98 & 68.66$ \pm $0.69 & 50.91$ \pm $2.57 & 83.12$ \pm $0.80 & 67.43$ \pm $0.81 \\
                Transformer-XH & 45.28$ \pm $1.08 & 38.39$ \pm $0.80 & 67.81$ \pm $0.93 & 76.51$ \pm $2.09 & 72.01$ \pm $0.86 & 56.26$ \pm $0.64 & 85.44$ \pm $0.75 & 68.13$ \pm $0.52 \\
                Transformer-XH++ & 46.81$ \pm $1.52 & 41.06$ \pm $1.70 & 70.52$ \pm $0.55 & 78.49$ \pm $0.52 & 73.92$ \pm $0.58 & 57.82$ \pm $2.29 & 85.35$ \pm $0.45 & 69.76$ \pm $0.73 \\
                MultiVerS & 51.56$ \pm $1.30 & 38.71$ \pm $1.96 & 66.32$ \pm $1.27 & 70.01$ \pm $2.23 & 81.33$ \pm $1.63 & \textbf{62.30}$ \pm $\textbf{0.98} & 78.14$ \pm $1.31 & 65.29$ \pm $0.36 \\
                CausalWalk & 45.52$ \pm $1.99 & 34.15$ \pm $0.97 & 71.49$ \pm $1.65 & 71.55$ \pm $2.46 & 71.27$ \pm $2.48 & 57.05$ \pm $0.62 & 80.65$ \pm $0.10 & 71.22$ \pm $1.74 \\
                \hline
                GPT2-PPL & 25.94$ \pm $1.00 & 25.58$ \pm $0.31 & 28.84$ \pm $0.14 & 29.00$ \pm $0.42 & 27.69$ \pm $1.56 & 30.35$ \pm $1.24 & 54.17$ \pm $0.05 & 54.14$ \pm $0.01 \\
                ProToCo & 42.63$ \pm $1.62 & 21.51$ \pm $1.22 & 36.68$ \pm $0.80 & 27.76$ \pm $1.35 & 52.94$ \pm $2.54 & 26.75$ \pm $0.91 & 40.12$ \pm $0.51 & 30.78$ \pm $0.85 \\
                ProgramFC & 46.04$ \pm $1.42 & 32.12$ \pm $0.76 & 62.49$ \pm $1.74 & 71.63$ \pm $0.91 & 60.17$ \pm $3.34 & 53.67$ \pm $1.92 & 86.84$ \pm $0.84 & 69.41$ \pm $2.07 \\
                \hline
                P-Tuning v2 & 52.54$ \pm $0.55 & 36.94$ \pm $0.13 & 73.03$ \pm $1.76 & 75.60$ \pm $3.01 & 76.56$ \pm $1.77 & 55.48$ \pm $2.04 & 87.01$ \pm $0.36 & 68.87$ \pm $0.76 \\
                \hline
                JustiLM & 47.33$ \pm $3.81 & 33.27$ \pm $1.98 & 58.75$ \pm $3.08 & 60.03$ \pm $1.60 & 69.63$ \pm $1.53 & 51.78$ \pm $0.80 & 81.33$ \pm $1.97 & 65.49$ \pm $0.65 \\
                \hline
                CORRECT & \textbf{59.88}$ \pm $\textbf{2.03} & \textbf{44.25}$ \pm $\textbf{1.73} & \textbf{75.34}$ \pm $\textbf{1.02} & \textbf{80.59}$ \pm $\textbf{1.00} & \textbf{83.20}$ \pm $\textbf{0.80} & 60.26$ \pm $1.31 & \textbf{88.41}$ \pm $\textbf{0.19} & \textbf{74.95}$ \pm $\textbf{0.38} \\
			\bottomrule
		\end{tabular}
	}
	\label{table:fully_supervised_macro_f1}
\end{table*}

\begin{table*}[t]
	\centering
	\caption{Verdict prediction results on \emph{fully supervised} setting with \emph{Micro F1} score. Results are in percentage.}
	\vspace{-0.2cm}
	\resizebox{\textwidth}{!}{
		\begin{tabular}{c|cc|cc|cc|cc}
			\toprule
                \multirow{2}{*}{Model} & \multicolumn{2}{c|}{BearFact} & \multicolumn{2}{c|}{Check-COVID} & \multicolumn{2}{c|}{SciFact} & \multicolumn{2}{c}{FEVEROUS-S} \\
			\cline{2-9}
			{} & Gold & Retrieved & Gold & Retrieved & Gold & Retrieved & Gold & Retrieved \\
			\hline
                KGAT & 69.42$ \pm $0.87 & 57.36$ \pm $0.52 & 72.05$ \pm $1.58 & 76.47$ \pm $0.65 & 74.44$ \pm $0.96 & 62.33$ \pm $0.88 & 86.21$ \pm $0.28 & 67.99$ \pm $0.78 \\
                HESM & 63.68$ \pm $1.39 & 58.62$ \pm $0.32 & 63.47$ \pm $0.25 & 71.90$ \pm $1.85 & 72.44$ \pm $0.77 & 53.36$ \pm $2.33 & 83.30$ \pm $0.75 & 68.36$ \pm $0.86 \\
                Transformer-XH & 61.26$ \pm $0.72 & 56.55$ \pm $1.50 & 68.56$ \pm $0.87 & 76.91$ \pm $1.51 & 75.89$ \pm $0.51 & 58.67$ \pm $1.53 & 85.61$ \pm $0.80 & 69.78$ \pm $0.41 \\
                Transformer-XH++ & 64.02$ \pm $1.44 & 58.39$ \pm $1.11 & 70.60$ \pm $0.67 & 78.65$ \pm $0.38 & 77.78$ \pm $0.69 & 60.56$ \pm $1.83 & 85.52$ \pm $0.39 & 70.37$ \pm $0.77 \\
                MultiVerS & 62.93$ \pm $1.17 & 50.69$ \pm $1.46 & 66.65$ \pm $1.71 & 70.70$ \pm $1.73 & 83.68$ \pm $1.40 & \textbf{66.77}$ \pm $\textbf{0.14} & 83.57$ \pm $1.54 & 67.66$ \pm $1.65 \\
                CausalWalk & 69.31$ \pm $1.69 & 60.00$ \pm $0.69 & 71.86$ \pm $1.54 & 71.68$ \pm $2.48 & 77.34$ \pm $2.30 & 59.00$ \pm $1.20 & 86.42$ \pm $0.92 & 71.51$ \pm $1.66 \\
                \hline
                GPT2-PPL & 40.00$ \pm $2.43 & 39.49$ \pm $0.73 & 32.75$ \pm $0.62 & 32.54$ \pm $0.93 & 31.50$ \pm $0.71 & 31.24$ \pm $1.56 & 54.33$ \pm $0.06 & 54.23$ \pm $0.01 \\
                ProToCo & 56.03$ \pm $0.24 & 35.57$ \pm $2.35 & 37.12$ \pm $1.10 & 32.75$ \pm $2.31 & 60.00$ \pm $1.53 & 31.17$ \pm $0.71 & 54.21$ \pm $0.67 & 44.71$ \pm $1.95 \\
                ProgramFC & 62.00$ \pm $2.83 & 54.63$ \pm $3.66 & 65.50$ \pm $2.12 & 72.36$ \pm $1.85 & 65.40$ \pm $3.78 & 59.76$ \pm $3.54 & 86.48$ \pm $0.33 & 69.50$ \pm $2.12 \\
                \hline
                P-Tuning v2 & 70.69$ \pm $0.49 & 60.34$ \pm $0.15 & 72.93$ \pm $1.85 & 77.32$ \pm $1.96 & 80.34$ \pm $0.94 & 57.44$ \pm $1.83 & 87.16$ \pm $0.33 & 70.56$ \pm $0.54 \\
                \hline
                JustiLM & 62.41$ \pm $1.25 & 48.49$ \pm $2.21 & 59.71$ \pm $2.56 & 61.71$ \pm $2.05 & 72.20$ \pm $1.85 & 54.74$ \pm $0.82 & 81.60$ \pm $0.33 & 68.38$ \pm $1.45 \\
                \hline
                CORRECT & \textbf{74.60}$ \pm $\textbf{1.11} & \textbf{61.84}$ \pm $\textbf{0.11} & \textbf{75.33}$ \pm $\textbf{0.93} & \textbf{80.83}$ \pm $\textbf{0.76} & \textbf{85.17}$ \pm $\textbf{0.71} & 63.50$ \pm $1.17 & \textbf{88.51}$ \pm $\textbf{0.19} & \textbf{75.35}$ \pm $\textbf{0.28} \\
			\bottomrule
		\end{tabular}
	}
	\label{table:fully_supervised_micro_f1}
\end{table*}

\textbf{Baselines.} We have 4 categories of baselines. 

\emph{\textbf{i}}) \textbf{Multi-hop fact-checking}, KGAT \cite{kgat}, HESM \cite{hesm}, Transformer-XH \cite{transformer_xh}, MultiVerS \cite{multivers}, and the recent CausalWalk \cite{causalwalk}. MultiVerS models contextual documents, and HESM and Transformer-XH incorporate referential documents. By comparing to them, we highlight the advantage of three-layer graph for modeling both contextual and referential documents. Since our model is built on Transformer-XH, we further extend it by modeling both contextual and referential documents, and name it Transformer-XH++. The comparison showcases the effect of evidence-conditioned prompting.

\emph{\textbf{ii}}) \textbf{Few-shot fact-checking}, GPT2-PPL \cite{gpt2_ppl}, ProToCo \cite{protoco}, and ProgramFC \cite{programfc}. They are mainly designed for few-shot setting. By increasing their training set, we could also compare to them on fully supervised setting. ProToCo and ProgramFC are proposed with handcrafted natural language prompt. By comparison, we verify the usefulness of our evidence-conditioned prompt embedding.

\emph{\textbf{iii}}) \textbf{Prompt tuning} is not for fact-checking. But for completeness, we convert P-Tuning v2 \cite{p_tuning_v2}, a continuous prompting, to our task.

\emph{\textbf{iv}}) \textbf{Retrieval-augmented generation for fact-checking}. Though our model is not designed with retrieval-augmented generation, we still compare to JustiLM \cite{justilm} for completeness.

\textbf{Implementation details.} Following \cite{transformer}, we set $ L $ to 12 and $ d $ to 768. Number of prompt embeddings $ M $ is 8. Temperature $ \tau $ in Eq. \ref{eq:scaling_and_shifting} is 100. For both our model and language model-based baselines, we initialize the model with pre-trained parameters in biomedical domain \cite{pubmedbert} for scientific datasets, and in general domain \cite{bert} for FEVEROUS-S. Each result is obtained by 5 independent runs. Experiments are done on 4 NVIDIA A100 80GB GPUs. More details are in Appendix \ref{sec:experiment_environment}.

We present two experimental settings below.

\begin{table*}[t]
	\centering
	\caption{Verdict prediction results on \emph{5-shot} setting with \emph{Macro F1} score. Results are in percentage.}
	\resizebox{\textwidth}{!}{
		\begin{tabular}{c|cc|cc|cc|cc}
			\toprule
                \multirow{2}{*}{Model} & \multicolumn{2}{c|}{BearFact} & \multicolumn{2}{c|}{Check-COVID} & \multicolumn{2}{c|}{SciFact} & \multicolumn{2}{c}{FEVEROUS-S} \\
			\cline{2-9}
			{} & Gold & Retrieved & Gold & Retrieved & Gold & Retrieved & Gold & Retrieved \\
			\hline
                KGAT & 36.62$ \pm $2.28 & 29.92$ \pm $3.99 & 35.65$ \pm $4.56 & 47.81$ \pm $2.40 & 39.07$ \pm $2.06 & 35.12$ \pm $2.79 & 50.21$ \pm $0.95 & 50.68$ \pm $1.21 \\
                HESM & 35.40$ \pm $3.77 & 26.00$ \pm $2.34 & 35.41$ \pm $4.78 & 42.82$ \pm $6.50 & 38.87$ \pm $1.69 & 34.03$ \pm $5.61 & 51.36$ \pm $0.35 & 51.92$ \pm $0.39 \\
                Transformer-XH & 29.45$ \pm $2.49 & 31.69$ \pm $2.06 & 40.48$ \pm $2.73 & 49.24$ \pm $1.60 & 47.65$ \pm $3.99 & 33.47$ \pm $1.11 & 52.45$ \pm $2.71 & 49.41$ \pm $1.93 \\
                Transformer-XH++ & 31.34$ \pm $4.07 & 29.74$ \pm $1.30 & 38.73$ \pm $1.35 & 50.56$ \pm $0.64 & 47.53$ \pm $0.65 & 33.79$ \pm $1.87 & 58.19$ \pm $0.75 & 52.78$ \pm $1.60 \\
                MultiVerS & 24.34$ \pm $3.12 & 20.92$ \pm $0.48 & 32.16$ \pm $2.50 & 50.80$ \pm $1.78 & \textbf{52.29}$ \pm $\textbf{1.92} & 29.64$ \pm $1.53 & 38.26$ \pm $0.18 & 38.82$ \pm $0.37 \\
                CausalWalk & 32.01$ \pm $3.35 & 31.10$ \pm $1.56 & 31.73$ \pm $5.13 & 43.79$ \pm $3.25 & 39.48$ \pm $5.51 & 34.95$ \pm $5.28 & 59.46$ \pm $1.78 & 55.37$ \pm $5.55 \\
                \hline
                GPT2-PPL & 24.99$ \pm $0.90 & 26.28$ \pm $0.18 & 25.05$ \pm $4.47 & 23.89$ \pm $2.65 & 27.69$ \pm $0.41 & 27.45$ \pm $0.66 & 51.33$ \pm $2.55 & 51.54$ \pm $2.50 \\
                ProToCo & 35.11$ \pm $0.40 & 21.51$ \pm $0.78 & 35.62$ \pm $5.32 & 29.72$ \pm $3.85 & 48.68$ \pm $3.38 & 25.93$ \pm $5.60 & 40.48$ \pm $0.88 & 31.00$ \pm $0.57 \\
                ProgramFC & 31.42$ \pm $1.20 & 30.88$ \pm $1.98 & 36.17$ \pm $0.73 & 49.06$ \pm $1.14 & 48.69$ \pm $0.46 & 33.18$ \pm $0.89 & 49.13$ \pm $2.57 & 51.62$ \pm $0.62 \\
                \hline
                P-Tuning v2 & 35.68$ \pm $2.36 & 31.86$ \pm $0.33 & 38.90$ \pm $4.81 & 50.63$ \pm $4.22 & 43.94$ \pm $0.54 & 33.33$ \pm $2.48 & 56.70$ \pm $1.82 & 48.53$ \pm $2.23 \\
                \hline
                JustiLM & 31.38$ \pm $2.07 & 26.01$ \pm $2.08 & 36.48$ \pm $2.78 & 44.39$ \pm $2.41 & 44.42$ \pm $2.08 & 31.04$ \pm $1.47 & 45.35$ \pm $1.18 & 42.48$ \pm $1.02 \\
                \hline
                CORRECT & \textbf{40.91}$ \pm $\textbf{1.42} & \textbf{33.47}$ \pm $\textbf{0.46} & \textbf{40.77}$ \pm $\textbf{1.19} & \textbf{52.40}$ \pm $\textbf{1.21} & 49.12$ \pm $0.30 & \textbf{35.30}$ \pm $\textbf{1.05} & \textbf{61.00}$ \pm $\textbf{1.95} & \textbf{57.04}$ \pm $\textbf{0.68} \\
			\bottomrule
		\end{tabular}
	}
	\label{table:5_shot_macro_f1}
\end{table*}
\begin{table*}[t]
	\centering
	\caption{Verdict prediction results on \emph{5-shot} setting with \emph{Micro F1} score. Results are in percentage.}
	\resizebox{\textwidth}{!}{
		\begin{tabular}{c|cc|cc|cc|cc}
			\toprule
                \multirow{2}{*}{Model} & \multicolumn{2}{c|}{BearFact} & \multicolumn{2}{c|}{Check-COVID} & \multicolumn{2}{c|}{SciFact} & \multicolumn{2}{c}{FEVEROUS-S} \\
			\cline{2-9}
			{} & Gold & Retrieved & Gold & Retrieved & Gold & Retrieved & Gold & Retrieved \\
			\hline
                KGAT & 44.66$ \pm $0.25 & 36.78$ \pm $3.86 & 37.55$ \pm $2.86 & 50.98$ \pm $1.13 & 42.22$ \pm $3.06 & 36.78$ \pm $3.52 & 51.13$ \pm $1.55 & 51.66$ \pm $1.38 \\
                HESM & 48.85$ \pm $2.42 & 28.97$ \pm $3.26 & 36.68$ \pm $5.15 & 50.11$ \pm $3.22 & 39.56$ \pm $2.45 & 35.89$ \pm $4.67 & 56.33$ \pm $0.91 & 53.68$ \pm $0.37 \\
                Transformer-XH & 32.53$ \pm $3.13 & 40.80$ \pm $3.32 & 41.67$ \pm $2.19 & 51.63$ \pm $1.16 & 48.89$ \pm $2.84 & 35.11$ \pm $1.95 & 52.94$ \pm $2.65 & 51.56$ \pm $0.73 \\
                Transformer-XH++ & 37.93$ \pm $4.10 & 35.06$ \pm $3.66 & 41.40$ \pm $1.78 & 52.41$ \pm $1.22 & 50.00$ \pm $0.85 & 36.67$ \pm $1.65 & 59.97$ \pm $1.50 & 53.23$ \pm $1.19 \\
                MultiVerS & 40.86$ \pm $0.74 & 39.49$ \pm $1.70 & 41.01$ \pm $1.29 & 49.82$ \pm $1.56 & \textbf{54.99}$ \pm $\textbf{1.90} & 43.33$ \pm $1.74 & 51.39$ \pm $1.33 & 51.84$ \pm $1.54 \\
                CausalWalk & 45.52$ \pm $3.47 & 41.38$ \pm $3.24 & 37.70$ \pm $4.59 & 43.79$ \pm $3.25 & 44.02$ \pm $2.70 & 41.78$ \pm $2.71 & 60.60$ \pm $1.98 & 55.20$ \pm $4.18 \\
                \hline
                GPT2-PPL & 36.38$ \pm $4.14 & 40.69$ \pm $0.49 & 33.19$ \pm $0.67 & 34.62$ \pm $0.72 & 29.44$ \pm $2.50 & 29.00$ \pm $1.46 & 53.02$ \pm $1.11 & 53.11$ \pm $1.12 \\
                ProToCo & 51.03$ \pm $2.44 & 33.80$ \pm $2.44 & 41.05$ \pm $2.47 & 34.50$ \pm $2.31 & 51.55$ \pm $2.27 & 36.78$ \pm $1.35 & 54.72$ \pm $1.32 & 42.45$ \pm $2.73 \\
                ProgramFC & 38.45$ \pm $2.19 & 38.28$ \pm $0.49 & 37.55$ \pm $0.38 & 50.00$ \pm $1.08 & 49.93$ \pm $0.36 & 36.17$ \pm $1.25 & 52.94$ \pm $2.67 & 51.94$ \pm $0.25 \\
                \hline
                P-Tuning v2 & 48.70$ \pm $1.95 & 41.90$ \pm $3.10 & 41.05$ \pm $3.88 & 52.07$ \pm $3.09 & 45.78$ \pm $1.07 & 37.67$ \pm $1.85 & 58.02$ \pm $1.68 & 52.06$ \pm $0.76 \\
                \hline
                JustiLM & 42.70$ \pm $1.64 & 37.72$ \pm $5.08 & 40.82$ \pm $2.20 & 46.73$ \pm $1.53 & 47.41$ \pm $1.07 & 33.00$ \pm $1.58 & 52.54$ \pm $1.71 & 49.38$ \pm $1.60 \\
                \hline
                CORRECT & \textbf{51.72}$ \pm $\textbf{1.04} & \textbf{42.76}$ \pm $\textbf{0.73} & \textbf{43.37}$ \pm $\textbf{1.82} & \textbf{54.46}$ \pm $\textbf{0.76} & 53.00$ \pm $1.30 & \textbf{44.36}$ \pm $\textbf{0.84} & \textbf{63.33}$ \pm $\textbf{0.91} & \textbf{57.14}$ \pm $\textbf{0.82} \\
			\bottomrule
		\end{tabular}
	}
	\label{table:5_shot_micro_f1}
\end{table*}
\begin{figure*}[h]
	\centering
	\includegraphics[width=1\textwidth]{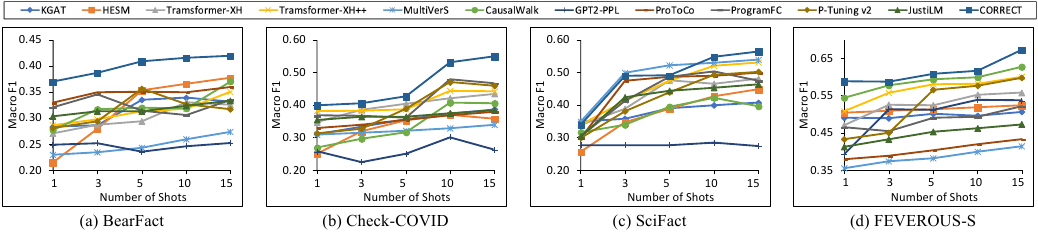}
	\caption{Few-shot veracity prediction with different number of shots.}
	\label{fig:different_shots_results}
\end{figure*}

\textbf{Fully supervised v.s. Few-shot.} For fully supervised setting, we train the model on the training set. If the dataset provides data split, we follow the split and obtain training and test sets. Otherwise, we split the dataset into 80:20 for training and test, respectively. Among training set, we further reserve 10\% for validation. For few-shot setting, we report 5-shot experiments as the main results, i.e., for each class in the label set $ y\in\mathcal{Y} $, we randomly sample 5 instances from training set, obtaining $ 5\times|\mathcal{Y}| $ training instances. This setting is consistent with existing work \cite{protoco}. For a fair comparison, we sample instances using 5 random seeds. We keep the same sampling for our model and baselines. We report the results on test set.

\textbf{Gold v.s. Retrieved evidence.} For gold evidence setting, we observe the ground-truth evidence sentences, and we verify the claim based on the gold sentences. For retrieved evidence setting, we do not observe any evidence sentences, and retrieve sentences from an evidence corpus, based on which we make prediction. We follow \cite{programfc} and use BM25 \cite{bm25} to retrieve top-3 evidence sentences for each claim. In the original Check-COVID dataset, if a claim is labeled as REFUTE based on the evidence, this claim is \emph{reused} in NEI class with another random evidence. Thus, there are two claims with the same content, but different evidence and labels. However, in our retrieved evidence setting, both claims will receive the same retrieved evidence, but they are labeled differently, making model training inconsistent. Thus, for retrieved evidence setting, we remove claims in NEI class for Check-COVID.

\subsection{Empirical Evaluation}

\textbf{Fully supervised setting.} We follow \cite{multivers} and report Macro F1 score for both gold and retrieved evidence settings in Table \ref{table:fully_supervised_macro_f1}. We also show Micro F1 score in Table \ref{table:fully_supervised_micro_f1}. Transformer-XH++ consistently outperforms Transformer-XH, verifying that contextual and referential documents bring useful information. By comparing CORRECT to Transformer-XH++, we design evidence-conditioned prompting to integrate evidence and claim embeddings, and further improve the performance. Models with handcrafted prompt do not predict verdict as accurately as our model, which showcases the advantage of continuous prompt embeddings. Overall, the results on gold evidence setting are higher than on retrieved evidence setting, because the retrieved evidence sentences may not be always correct and may contain noisy information. The only exception is Check-COVID, because the retrieved evidence setting has only two labels, making the prediction task easier. MultiVerS is slightly better than CORRECT on SciFact, because the evidence sentences in SciFact contain sufficient information for fact-checking as shown in \cite{scifact,multivers}, and referential documents do not bring much additional benefit.

\begin{figure*}[t]
	\centering
	\includegraphics[width=1\textwidth]{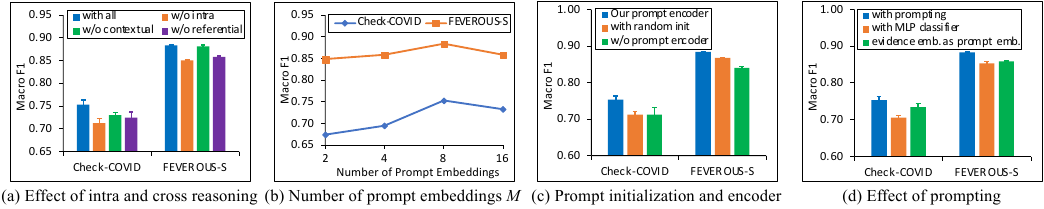}
	\caption{Model analysis on Check-COVID and FEVEROUS-S.}
	\label{fig:model_analysis}
\end{figure*}

\begin{figure}[t]
	\centering
	\includegraphics[width=0.8\linewidth]{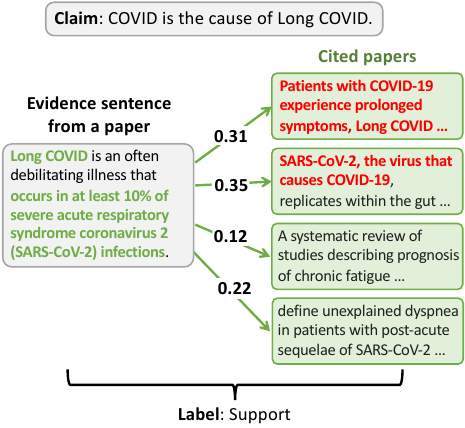}
	\caption{Case study on BearFact dataset.}
	\label{fig:case_study}
\end{figure}

\textbf{Few-shot setting.} We report 5-shot results in Table \ref{table:5_shot_macro_f1} for Macro F1 score and Table \ref{table:5_shot_micro_f1} for Micro F1 score. Overall, HESM and Transformer-XH perform better than others, since referential documents contain useful information to complement evidence sentences for accurate prediction. Our model further improves them, verifying the strength of both contextual and referential documents. P-Tuning v2 outperforms models with handcrafted prompt, since continuous prompt embeddings can better adapt to the training data. By comparing to it, we design an evidence-conditioned prompt encoder to integrate contextual and referential documents into prompt embeddings, and produce more accurate results. We vary the number of shots in $ \{1, 3, 5, 10, 15\} $ in Fig. \ref{fig:different_shots_results}. Though our model is competitive with MultiVerS on SciFact, we are still better than it on other datasets, due to the advantage of both contextual and referential documents.

\subsection{Model Analysis}

\textbf{Effect of intra- and cross-layer reasoning.} We respectively remove each graph layer from the complete model. Macro F1 score is shown in Fig. \ref{fig:model_analysis}(a). The model with all three layers performs the best, indicating that all three layers bring useful information. Intra-layer reasoning on evidence sentence layer plays the most important role, since evidence sentences provide the most immediate information for verification. Contexts and references are important, and disregarding them deteriorates the results.

\textbf{Different numbers of prompt embeddings $ M $.} We vary the number of prompt embeddings $ M $ in Fig. \ref{fig:model_analysis}(b). When $ M=2 $, we cannot fully capture the interaction between evidence and claims, causing a low accuracy. After we increase $ M $, we observe an improvement. An overly high $ M $ hurts the result, because overfitting problem appears.

\textbf{Prompt initialization and encoder.} Our prompt encoder has both initialization of base prompt embeddings and evidence-conditioned prompt encoder. \emph{\textbf{i}}) To test the effect of initialization, we replace it with random initialization and report the results in Fig. \ref{fig:model_analysis}(c). Our initialization produces better results, because evidence graph separates different sets of prompt embeddings and provides a more informative starting point. \emph{\textbf{ii}}) We remove prompt encoder from our model, and only retain base prompt embeddings. Fig. \ref{fig:model_analysis}(c) shows that removing prompt encoder hurts the results, indicating that prompt encoder is necessary to combine evidence and claim for accurate prediction.

\textbf{Effect of prompting.} We design two ablated models. \emph{\textbf{i}}) We replace prompting with an MLP classifier, which concatenates evidence and claim embeddings as input, and produces predicted label. Here claim embedding is obtained without prompt embeddings. \emph{\textbf{ii}}) We directly consider evidence embedding as prompt embedding, and do not assume base prompt embeddings. Fig. \ref{fig:model_analysis}(d) shows that prompting performs better than MLP classifier, because prompt embeddings and claim token embeddings are input to claim encoder together, and the contextualized encoding helps exchange information between evidence and claim for accurate prediction. Base prompt embeddings are also helpful, since they store general fact-checking knowledge and help generalize across different claims.

\textbf{Case study.} To intuitively understand that our model captures useful information in referential documents, we conduct a case study and visually show the attention values between an evidence sentence and its cited papers in graph neural networks. Fig. \ref{fig:case_study} shows that the highest attention scores appear between the evidence sentence and referential documents that indeed contain useful information. This visualization verifies that referential documents are crucial to improve claim verification.
\section{Conclusion}

We propose a context- and reference-augmented reasoning and prompting model for fact-checking. To model contextual and referential documents, we construct a three-layer graph with intra- and cross-layer reasoning. To integrate evidence into claims, we design evidence-conditioned prompting, which produces unique prompt embeddings for each claim. A future work is to extend three-layer graph to a multi-modal graph for fact-checking.
\section*{Acknowledgments}

This work was in part supported by NSF awards \#1934782 and \#2114824. Some of the research results were obtained using computational resources provided by NAIRR award \#240336.
\clearpage
\section*{Limitations}

Here we identify two limitations of our work in terms of dataset and evidence type.

\textbf{Dataset.} Our model is proposed to incorporate contextual and referential documents of evidence sentences. We assume that the contextual and referential documents of evidence sentences are available in the dataset, or the dataset provides identifiers for evidence sentences, such as PubMed ID, so that we can use these identifiers to search their contextual and referential documents online. In Appendix \ref{sec:dataset_preprocessing}, we provide details on how to use identifiers to obtain contextual and referential documents. If the given dataset does not provide contextual or referential documents, or the identifiers of evidence sentences are not available, our model will reason within evidence sentences for fact-checking.

\textbf{Evidence type.} Following existing textual fact-checking models, we propose our model to reason over textual evidence sentences only. Our model is not proposed for tabular or multi-modal evidence, thus cannot reason over these types of evidence for fact-checking. One potential future work would be to extend our three-layer evidence graph to a multi-modal graph for evidence reasoning.
\section*{Ethics Statement}

We do not foresee any undesired implications stemming from our work. Conversely, we hope that our work can advance AI Ethics research.

\bibliography{acl_latex}

\appendix
\section{Pseudo-code of Training Process}
\label{sec:algorithm}

We summarize the training process at Algo. \ref{algo:training_algorithm}.

\begin{algorithm}[h]
	\caption{Training Process of CORRECT}
	\label{algo:training_algorithm}
	\begin{flushleft}
		\hspace*{\algorithmicindent}\textbf{Input}: A fact-checking dataset $ \mathcal{D} $ with claims $ \mathcal{X} $, evidence sentences $ \mathcal{E} $, contextual documents $ \mathcal{C} $, and referential documents $ \mathcal{R} $. Number of prompt embeddings $ M $ and temperature $ \tau $. \\
		\hspace*{\algorithmicindent}\textbf{Output}: Predicted labels $ \widehat{\mathcal{Y}}_{\text{test}} $ for test claims.
	\end{flushleft} 
	\begin{algorithmic}[1] 
		\State Initialize model with pre-trained parameters in biomedical domain or general domain.
		\While{not converged}
        \State Construct three-layer evidence graph for each claim $ x\in\mathcal{X} $.
            \For{evidence sentence $ e\in\mathcal{N}_{\text{evid}}(x) $}
            \State Initialization $ \textbf{H}_e^{(l=1)}=\text{TRM}(\textbf{H}_e^{(l=0)}) $.
                \For{$ l=1,2,...,L-1 $}
                \Statex \qquad\qquad\textit{ // Evidence graph reasoning}
                \State Intra-layer reasoning by Eqs. \ref{eq:linear_projection}--\ref{eq:graph_conv_layer}.
                \State Cross-layer reasoning by Eq. \ref{eq:graph_conv_layer_ref}.
                \Statex \qquad\qquad\textit{ // Asymmetric MHA step}
                \State Virtual token concatenation Eq. \ref{eq:virtual_token}.
                \State $ \textbf{H}_e^{(l+1)}=\text{TRM}^{\text{asy}}(\widehat{\textbf{H}}_e^{(l)}) $ by Eq. \ref{eq:asymmetric_attention}.
                \EndFor
            \EndFor
        \State Obtain an evidence embedding by Eq. \ref{eq:evidence_embedding}.
        \Statex \quad\textit{ // Evidence-conditioned prompting}
        \State Initialize $ |\mathcal{Y}| $ sets of base prompt embeddings $ \{\textbf{h}_{m,y}\}_{m=1}^M $ where $ y\in\mathcal{Y} $.
        \State Input evidence embedding $ \textbf{h}_E $ to evidence-conditioned prompt encoder and obtain $ |\mathcal{Y}| $ sets of prompt embeddings $ \{\bm{\pi}_{m,y}\}_{m=1}^M $ where $ y\in\mathcal{Y} $ by Eqs. \ref{eq:scaling_and_shifting}--\ref{eq:scaling_and_shifting2}.
        \State Input $ \{\textbf{P}_{x,y}\}_{y\in\mathcal{Y}} $ to claim encoder and obtain $ |\mathcal{Y}| $ claim embeddings $ \{\textbf{h}_{x,y}\}_{y\in\mathcal{Y}} $.
        \State Minimize loss function $ \mathcal{L} $ in Eq. \ref{eq:objective_function}.
    \EndWhile
	\end{algorithmic}
\end{algorithm}

\section{Dataset Preprocessing Details}
\label{sec:dataset_preprocessing}

Here we present details of dataset preprocessing.

\textbf{FEVEROUS}\footnote{\url{https://fever.ai/dataset/feverous.html}} \cite{feverous} is a general-domain dataset, and each claim is annotated in the form of sentences and/or cells from tables in Wikipedia pages. In this paper we mainly focus on textual evidence sentences, thus we follow ProgramFC \cite{programfc} and obtain claims that only require textual evidence for verification, and name this subset \textbf{FEVEROUS-S}. Claims in this dataset have two labels only, SUPPORT and REFUTE. In the original dataset, each evidence sentence may contain hyperlinks to other Wikipedia pages, and such hyperlinks in sentences are indicated with double square brackets. We thus retrieve words or phrases inside double square brackets, and use them as entries to query Wikipedia dump to obtain their corresponding pages as referential documents. FEVEROUS uses the December 2020 dump, including 5.4 million full Wikipedia articles. If a Wikipedia page has overly many sentences, we reserve its top-20 sentences, since almost all the evidence sentences appear within top-20 sentences in FEVEROUS. Similarly, the full content of the Wikipedia page is contextual document of each evidence sentence. If a page has overly long content, we reserve its top-20 sentences.

\textbf{BearFact}\footnote{\url{https://www.ims.uni-stuttgart.de/en/research/resources/corpora/bioclaim/}} \cite{bear_fact} is a biomedical claim verification dataset. Evidence sentences are obtained from paper abstracts in PubMed database\footnote{\url{https://pubmed.ncbi.nlm.nih.gov/}}. Original dataset does not provide evidence sentences for claims in NEI class. Thus we follow existing work \cite{protoco} and select evidence sentences that have the highest \textit{tf-idf} similarity with claims as their evidence. We consider the full abstract as the contextual document for each evidence sentence, as in MultiVerS \cite{multivers}. In addition, we use S2ORC \cite{s2orc} to obtain cited papers with abstracts as referential documents. Specifically, the original dataset provides PubMed ID for each evidence sentence. We use PubMed IDs as identifiers to search in S2ORC database and obtain cited papers. If a paper has overly many citations, we reserve its top-20 citations to avoid data redundancy.

\textbf{Check-COVID}\footnote{\url{https://github.com/posuer/Check-COVID/tree/main/Check-COVID}} specifically focuses on COVID-19 claims taken from news articles. Each evidence sentence is from a paper abstract with CORD ID as identifier. We thus use CORD IDs to search in S2ORC database and obtain cited papers. Similarly, we consider the full abstract as contextual document. The original dataset provides sentences for claims in NEI class.

\textbf{SciFact}\footnote{\url{https://github.com/allenai/scifact/tree/master}} \cite{scifact} is another biomedical fact-checking dataset with sentences in paper abstracts as evidence. Similarly, the full content of the abstract is considered as contextual document. In addition, each evidence sentence is coupled with S2ORC ID, which is used to obtain its citations using S2ORC database. The original dataset does not have sentences for claims in NEI class. Thus we follow the original paper \cite{scifact} and choose top-3 sentences in the same abstract with the highest \textit{tf-idf} similarity with the claim as evidence.
\section{Experiment Environment}
\label{sec:experiment_environment}

All the experiments were conducted on Linux server with 4 NVIDIA A100-SXM4-80GB GPUs. Its operating system is 20.04.5 LTS (Focal Fossa). We implemented our proposed model CORRECT using Python 3.9 as programming language and PyTorch 1.14.0 as deep learning library. Other frameworks include numpy 1.22.2, sklearn 0.24.2, and transformers 4.43.3.

\end{document}